\documentclass[conference]{IEEEtran}
\IEEEoverridecommandlockouts
\usepackage[numbers,sort,compress]{natbib}
\usepackage{xfrac}
\usepackage{booktabs}
\usepackage{url}
\usepackage{comment}
\usepackage{subfig}
\usepackage{amsmath,amssymb,amsfonts}
\usepackage{mathtools}
\usepackage{algorithmic}
\usepackage{graphicx}
\usepackage{textcomp}
\usepackage{xcolor}
\usepackage{nicefrac}
\def\BibTeX{{\rm B\kern-.05em{\sc i\kern-.025em b}\kern-.08em
    T\kern-.1667em\lower.7ex\hbox{E}\kern-.125emX}}
\usepackage{makecell}
\usepackage{pgfplots}
\pgfplotsset{compat=newest}
\usepackage{multicol}
\usepackage{afterpage}

\usepackage{xcolor}
\usepackage{tcolorbox}
\usepackage{tabularx}
\usepackage{float}
\definecolor{navyblue}{rgb}{0.0, 0.0, 0.5}
\usepackage{color}
\usepackage[colorlinks=true,linkcolor= {red!50!black}, urlcolor=navyblue, citecolor=navyblue, pdfborder={2 1 0}]{hyperref}
\newcommand{\MYhref}[3][navyblue]{\href{#2}{\color{#1}{#3}}}%
\usepackage[font=scriptsize,labelfont=bf]{caption}

\begin{document}

\newtheorem{theorem}{Theorem}
\newtheorem{lemma}{Lemma}
\newtheorem{definition}{Definition}

\renewcommand{\vec}[1]{\mathbf{#1}}
\newcommand{\ve}[1]{\boldsymbol{#1}} 

\newcommand{\diag}{\mbox{diag}}
\newcommand{\tr}{\mbox{tr}}
\newcommand{\F}{{\cal{F}}}
\newcommand{\Om}{\Omega}
\newcommand{\om}{\omega}
\newcommand{\Pp}{ \mathbb{P} }
\newcommand{\R}{\mathbb{R}}
\newcommand{\Z}{\mathbb{Z}}
\newcommand{\E}{\mathbb{E}}
\newcommand{\En}{\mathbb{E}_n}
\newcommand{\Et}{\tilde{\mathbb{E}}}
\newcommand{\Etn}{\tilde{\mathbb{E}}_n}
\newcommand{\borel}{ {\cal{B}}(\mathbb{R}) }
\newcommand{\pspace}{(\Om, \F, \Pp)}
\newcommand{\Pt}{\tilde{\mathbb{P}}}
\newcommand{\T}{\intercal}
\newcommand{\Var}{\mbox{Var}}
\newcommand{\Cov}{\mbox{Cov}}
\newcommand{\RR}{\varrho _{1,2}}

\newcommand\nd{\textsuperscript{nd}\xspace}
\newcommand\rd{\textsuperscript{rd}\xspace}
\newcommand\nth{\textsuperscript{th}\xspace} 

\renewcommand{\footnoterule}{\vspace*{2cm}\noindent\hspace*{-1em}\smash{\rule[3pt]{0.4\columnwidth}{0.4pt}}}

\title{Variations on the Reinforcement Learning performance of Blackjack}

\author{
\IEEEauthorblockN{Avish Buramdoyal}
\IEEEauthorblockA{\textit{Department of Statistical Sciences} \\
\textit{University of Cape Town},\\
Rondebosch 7701, South Africa \\
orcid.org/0009-0004-3877-8704 \\
brmavi002@myuct.ac.za}
\and
\IEEEauthorblockN{Tim Gebbie}
\IEEEauthorblockA{\textit{Department of Statistical Sciences} \\
\textit{University of Cape Town},\\
Rondebosch 7701, South Africa \\
orcid.org/0000-0002-4061-2621 \\
tim.gebbie@uct.ac.za}
}

\maketitle

\begin{abstract}
Blackjack or ``21" is a popular card-based game of chance and skill. The objective of the game is to win by obtaining a 
hand total higher than the dealer’s without exceeding 21. The ideal blackjack strategy will maximize
financial return in the long run while avoiding gambler’s ruin. The stochastic environment and inherent reward structure of blackjack presents an appealing problem to better understand reinforcement learning agents in the presence of environment variations. Here we consider a q-learning solution for optimal play and investigate the rate of learning convergence of the algorithm as a function of deck size. A blackjack simulator allowing for universal blackjack rules is also implemented to demonstrate the extent to which a card counter perfectly using the basic strategy and hi-lo system can bring the house to bankruptcy and how environment variations impact this outcome. The novelty of our work is to place this conceptual understanding of the impact of deck size in the context of learning agent convergence. 
\end{abstract}
\vspace{0.1cm}

\begin{IEEEkeywords}
Blackjack, multi-agent dynamics, Q-learning
\end{IEEEkeywords}



\noindent

\section{Introduction} \label{sec:Introduction}

The origin of blackjack is still debated today. It seems that blackjack started around the 1700's in France \cite{blackjackorigin} and was called {\it vingt-et-un}, translating to 21 \cite{blackjackhistory1}. The game evolved extensively in North America where a house-banked blackjack system was first introduced in Nevada in 1931 \cite{blackjackhistory4}.  During the 20$^\text{th}$ century the game evolved to offer bonus payouts. This included one that paid an extra if a jack of spades or clubs {\it i.e.} a ``blackjack", was dealt together with an ace of spades \cite{blackjackorigin}. It was from then the game changed its name to blackjack \cite{blackjackhistory1}. The rise in prevalence of legalised games in the casinos of Las Vegas inspired a number of players to develop optimal strategies for play \cite{blackjackhistory1}.

Roger Baldwin, Wilbert Cantey, Herbert Maisel and James McDermott, known to blackjack insiders as the “Four Horse-men”, were the first in determining modern optimal blackjack strategies \cite{baileyinvented}. The basic strategy to play blackjack \cite{BaldwinOptimum} was published in 1956 and this spawned numerous attempts aimed at improvement. 

A seminal strategy success became widely known in the early 1960's when mathematician Edward Thorp published his book "Beat the dealer: a winning strategy for the game of Twenty-One" \cite{blackjackorigin}. This caught the attention of many players. Casinos then started to increase the deck size in an attempt to overcome the effectiveness of counting systems \cite{canyoustillbeat}. Despite this, blackjack is one of today's most popular gambling games, and is played in practically every casino worldwide. 

\section{Blackjack problem} \label{sec:Blackjack problem}

\subsection{Game setting} \label{sec:Game settings}
The objective of blackjack is to get a hand total higher than the dealer without ``busting". Busting occurs when the players hand exceeds a score of 21. Blackjack is a casino banked game allowing players to compete against the house rather than each other \cite{gamesettings1}. The game of blackjack consists of a dealer and 1 to 7 players. A standard deck of 52 cards was initially used, but after the announcement of the first winning strategies \cite{Ed_Thorp}, casinos implemented countermeasures. One countermeasure was to vary the deck size. Cards 2 to 10 are worth their face values. Jacks (J), queens (Q) and kings (K) are counted as 10 and an ace (A) is worth 1 or 11, whichever is the most favourable to the player. A hand with an ace valued at 11 is a ``soft hand", and all other hands are considered ``hard hands". The distinction between soft and hard hands is important as this defines a key difference in the strategy optimal to the player who is holding either hands. This distinction is understood by analysing the strategy tables provided in Section \ref{sec:Strategy Tables} of the Appendix.  

\subsection{Playing blackjack} \label{sec:Playing blackjack}

Before the start of any hand, each player places an initial bet based on their own bankroll and the minimum-maximum bet set by the casino. The players are then each dealt with two cards face up, and the dealer then also gets two cards; one face up and one face down -- the face down card is called a ``hole card". The dealing and drawing of cards occur in a clockwise fashion starting from the player left of the dealer \cite{gamesettings1}.

The player will look at his two cards and take actions requested from the dealer. After all players have finished their actions on a given round, the dealer then turns his face down card up and decides which action to take. The challenge posed to the player is the choice of optimal action at each hand, given his current hand total and the dealer's face up card. The possible actions at play include: standing, hitting, splitting, doubling down, surrendering and insurance.

\subsection{Available actions} \label{sec:Available actions}

A player can choose to ``stand", {\it i.e.} take no additional cards, if doing so is unattractive for his current hand total. Hitting is when a player requests another card from the dealer. A player is allowed to ``hit" as many times as he chooses, as long as he does not bust. A player can choose to separate two cards of the same face value and make another bet of equal size as the initial bet made and play each card as a separate hand. This is known as splitting. Here we do not consider the case where the player is allowed to re-split when they have cards of the same value after the first split. Another possible action to a player is to toss his cards away and give up half of his bet, i.e. surrender when he feels that he has a weak hand, based on his first two cards against the dealer's potential hand. A further choice available to the player is to double his bet in the middle of a hand. The player has to then draw only one card from the dealer. A further available choice is "insurance". Under insurance, an additional wager by the player is allowed on the condition that the dealer has an ace as the face up card, and the player does not have a blackjack \cite{canyoustillbeat}. The player expects the dealer to have a blackjack. If the dealer does have a blackjack, the player wins twice his side bet.

\subsection{Why not take insurance?} \label{sec:Why not take insurance?}

It has been recommended that a player should never take insurance \cite{insurance}. Using 6 packs of cards, making a total of 312 cards with indices +2 and -1 being assigned for winning and losing respectively, it is shown that irrespective of whether the player has a blackjack or not, and the dealer's face up is an ace, the expected winnings of the player is always less for taking insurance \cite{insurance}. We therefore do not incorporate insurance as a necessary allowable action in this study.

\subsection{Natural} \label{sec:Natural}
A starting hand of an ace and a 10-valued card is a ``blackjack" or ``natural" and beats any other hands -- even a hand total of 21 which is not a blackjack.

\subsection{Dealer stands or hits on soft 17} \label{sec:Dealer stands or hits on soft 17}
A ``soft 17" hand is one when the ace is counted as 11 and the hand total is 17 {\it e.g.} A-6, A-2-4, A-3-3. In the realm of blackjack, certain casinos require dealers to stand on a soft 17 hand while others necessitate dealers to hit on a soft 17 hand \cite{standorhit}. 

\subsection{Settlement} \label{sec:Settlement}
If the player and dealer have the same hand total, it is a ``tie" or ``draw", and no money is exchanged. If neither the player nor the dealer busts, the one with the higher and more favourable hand value wins the bet. If the player busts, the player loses the bet, irrespective of the dealer's score. Assuming blackjack pays 3:2, the player receives 1.5 times his initial bet for having a blackjack. For cases other than for a blackjack and insurance, the amount paid to the dealer by the player is always equal the total bet made by the player for that round.

\section{Basic strategy} \label{sec:Basic strategy}

For this project, we follow the blackjack "basic strategy" as initially determined by \citet{BaldwinOptimum}. This is a strategy maximizing a player's mathematical expectation given his strategic problem. A basic strategy is simply a proper playing decision for every possible hand against the dealer \cite{Ed_Thorp}. Every possible combination of the player hands' total and dealer's face up card has a mathematically correct play and these can be summarised in a strategy table framework adapted from \cite{mich}, as provided in Section \ref{sec:Strategy Tables} of the Appendix. The optimal action per round therefore depends on the player's current hand total and the dealer's face up card \cite{BaldwinOptimum}.

The player's basic decisions when playing blackjack are: when to draw or stand, when to double down, and when to split pairs \cite{BaldwinOptimum}. 

\section{Counting cards} \label{sec:Counting cards}

The optimal action per each round depends on the player’s current hand total and the dealer’s face up card \cite{Wong}. A good player can take advantage of the basic strategy by keeping track of cards dealt in previous plays \cite{Ed_Thorp}. The track, or sequence of plays realised at the table, is used as information for the next round and is known as ``card counting". Generally, a card counter looks for hands when more high cards are left to be played than a standard deck would have allowed \cite{intuition}. This was the key to Edward Thorp's Ten Count system which attempts to determine the ratio of high to low cards in a deck \cite{Ed_Thorp}.

\subsection{Ten count and Hi-lo system} \label{sec:Ten count and Hi-lo system}

In determining the ratio of high to low cards for a one-deck blackjack game, values are assigned to each card. The values for the Ten count system is given by Table \ref{tab:count1}. The player can now count cards. The player starts with a count of 0. Then, based on the indices, he will add or subtract for every single card revealed. This is known as the ``running count”.

\begin{table}[htbp]
\centering
\begin{tabular}{|c|c|}
\hline
\textbf{Card} & \textbf{Value Assigned}\\
\hline
A-9 & +4\\
\hline
10,K,Q,J & -9 \\
\hline
\end{tabular}
\caption{This table is the Ten count system proposed by \cite{Ed_Thorp}. This system assigns a value of +4 to cards (A-9) and a value of -9 to (10,J,K). The intuition the player follows is to then bet a higher proportion of his bankroll for higher counts and lower proportion for lower counts.}
\label{tab:count1}
\end{table}

The next step is to compute the ``true count”:

\begin{equation}
\text {true count }=\frac{\text {running count}}{\text {decks remaining }}.\label{eqn:1}
\end{equation}

The higher the count, the more the player should bet given his higher advantage. The general idea is to bet little or nothing when player advantage is low and to bet proportionately higher when player advantage is high \cite{Ed_Thorp}, {\it i.e.} in multiples of the player’s betting-unit\footnote{The size of player’s bet.} \cite{Ed_Thorp}. The mathematical theory of Edward Thorp behind sizing bets from a count was based on the Kelly criterion, an intermediate strategy between maximising one’s expected return growth rate while minimising the probability of ruin \cite{Thorpkelly}.

Harvey Dubner introduced a simplified variation of Thorp’s strategy, called the ``Hi-Lo card counting strategy” (or the ``point count system”), at the Fall Joint Computer Conference in Las Vegas in 1963 \cite{Patterson2002}. The Hi-Low system is the most commonly used card counting system nowadays \cite{mostcommon} and represented in Table \ref{tab:count2}.

\begin{table}[htbp]
\centering
\begin{tabular}{|c|c|}
\hline
\textbf{Card} & \textbf{Value Assigned}\\
\hline
2,3,4,5,6 & +1\\
\hline
7,8,9 & 0 \\
\hline
10,J,Q,K,A & -1\\
\hline
\end{tabular}
\caption{Under the Hi-Lo system, each card is assigned a value of either -1,+1 or 0. The lower cards (2-6) in the deck are given a value of +1. The ``neutral cards” (7-9) are given a zero value and the high cards (10,K,Q,J,A) are all given a -1 value. Using the true count, this system allows a player to know his edge and helps him to size his bet at the start of every round.}
\label{tab:count2}
\end{table}

The value of +1 means that as low cards are depleted from the game, chances of busting are low and so fewer cards can hurt you in the future while the value of -1 means that as high cards are depleted from the game, player advantage falls. The value of 0 represents neutral cards and favours neither the player nor the house.  

\subsection{Other counting systems} \label{sec:Other counting system}

In assessing the impact that a variation in deck size may have on the performance of a learning agent, we adapt the comparison made by \cite{counting_systems} on ``The Best Card Counting System: A Comparison of the Top 100" for blackjack. The methodology adopted by \cite{counting_systems} in searching for the ``best card counting system" involved ranking the different schemes by profit potential. 

The counting systems were tested for both single-deck and multi-deck games but ranged from level 1 to 4.  It has been found that higher level systems perform at a rate of profit of 0.1\% better than level one systems \cite{counting_systems}. We consider 2 other counting systems: Zen Count and Uston APC system. To use these two systems, the same counting principle is applied, meaning a player would still bet a higher proportion of his bankroll for having a high count and bet proportionately lower for lower counts. The Zen count framework is presented in Table \ref{tab:count3}.

\begin{table}[H]
\centering
\begin{tabular}{|c|c|}
\hline
\textbf{Card} & \textbf{Value Assigned}\\
\hline
2,3 & +1\\
\hline
4,5,6 & 2 \\
\hline
7 & 1\\
\hline
8,9 & 0\\
\hline
10, A & -2\\
\hline
\end{tabular}
\caption{For this counting method, the player uses these indices assigned and bets accordingly.}
\label{tab:count3}
\end{table}

The Zen count as suggested from a Blackjack formula used by \cite{counting_systems}, would yield a Rate of Profit of 2\% for a player for a single deck game, with a betting efficiency of 0.97.

The Uston APC system is a balanced count one, implying the running count will always end up 0 once the whole deck has been dealt out. Uston APC is the second best counting scheme \cite{counting_systems} and a player would use Table \ref{tab:count4} to count cards using the Uston APC method.

\begin{table}[H]
\centering
\begin{tabular}{|c|c|}
\hline
\textbf{Card} & \textbf{Value Assigned}\\
\hline
2,8 & +1\\
\hline
3,4,6,7 & 2 \\
\hline
5 & 3\\
\hline
9 & -1\\
\hline
10 & -3\\
\hline
A & 0\\
\hline
\end{tabular}
\caption{Uston APC method indices}
\label{tab:count4}
\end{table}

Compared to the Zen count, the Uston APC yields a Rate of Profit of 1.98\% for a player \cite{counting_systems}.

\section{Performance expectations} \label{sec:Performance expectations}
As suggested in ``Professional blackjack" by \cite{Wong}, a player's advantage or disadvantage varies with the rules and number of decks played. Commonly, a player's advantage is around 0.5\% with the basic strategy \cite{Wong}. 

\subsection{Deck size variation} \label{sec:Deck size variation}

It turns out that increasing the deck size only slightly cuts the player's advantage \cite{Ed_Thorp}. An analysis was conducted by \cite{conrad_kirk} using similar blackjack rules adapted from \cite{Ed_Thorp}. The analysis performed was to determine the player's disadvantage \% for using the basic strategy. The only difference in simulation was the absence of a cut card in the pack. It was found that a player could have an advantage for a single-deck game but player disadvantage increases when playing with more decks. In addition, it was also found that the fewer decks, the more blackjack you could win relative to a game with more decks \cite{easytowin}. 

\subsection{Bet size and strategy variation} \label{sec:Bet size and strategy variation}

\citet{Wong} suggests that a player's expected win \% is proportional to his bet size and the amount of time available for play. The more a player bets, the more he will win on lucky hands and the more he loses on unlucky hands. For every unit increase in the true count, given by Equation \ref{eqn:1}, player's advantage goes up by 0.5\% \cite{Wong}.

The variance of possible outcomes to the player depends on the specific rules set \cite{Wong}. For example, a player allowed to double down after splitting generates higher variance (bigger ups and downs). Alternatively, allowing a player to only double down on 10 or 11 means lower variability. Variance of outcomes also depends on the number of simultaneous hands being played \cite{Wong}.

\section{Reinforcement Learning} \label{sec:Reinforcement Learning}

\subsection{The Reinforcement Learning problem} \label{sec:The Reinforcement Learning problem}
Reinforcement learning (RL) is often considered to be a third machine learning paradigm alongside supervised and unsupervised learning \citep{Sutton}. RL problems involve relying on responses from an environment to learn \citep{Granville2005}, and more specifically to map situations to actions \citep{Sutton}. The responses under RL take the form of rewards to guide the agent in developing his policy \citep{Granville2005}. The aim is to maximise a numerical reward signal. 

The environment allows for interactions among agents. An agent is an entity making decisions in the environment and must have goal(s) relating to the environment. To encode these goals, we first need to specify how the agent behaves relative to states of the environment. A policy is used to define the agent's behaviour at a given time and is a mapping of states to actions when in those states \citep{Sutton}. A reward signal is then used to define the goal of the problem. At each time step, the agent is sent a single number, a reward. The agent's objective is to maximize his total rewards in the long run while avoiding gambler's ruin. The agent can influence the reward signal only through his actions and thus the reward sent to the agent at any given time depends on the agent's current action and state of the environment. Values advise the long-term worthiness of states after taking into account, the states to follow by the agent and the rewards attached to each of those states. In addition, the model mimics the behaviour of the environment allowing inferences to be drawn about the environment's future behaviour. 

For the blackjack case, the environment is the blackjack setup simulation allowing for play between the agent and dealer. The agent is the player playing against the dealer. An intuitive policy would be one allowing the player to get a score as close to 21 without busting. The reward to be maximized in the long run is the difference between the reward the agent receives or pays out at the end of each round. Value states here relates to the value to the agent of being in any particular state from the lookup table in Section \ref{sec:Strategy Tables} of the Appendix.

Blackjack can be formulated as an episodic finite Markov Decision Process (MDP) where each game is an episode \citep{Sutton}. Rewards +1, -1 and 0 are given for winning, losing and drawing respectively. We note that rewards $\gamma$ are not discounted, i.e. $\gamma = 1$ and so terminal rewards represent returns. The state depends on the player's hand total and the dealer's face up card. 

\subsection{Value functions} \label{sec:Value functions}

We aim to estimate ``how good" it might be for an agent to be in a given state, or to take an action in any particular state, {\it i.e.} ``how good" it might be for an agent to take an action given his hand total and the dealer's face up card. The estimates of ``how good" it might be for an agent to either be in a state or take an action is derived from experience using Monte Carlo methods \cite{Sutton}. 

\subsubsection{State-value function} \label{sec: State-value function}

The idea of ``how good" is captured through the state-value function, the expected returns of future rewards for the agent, and is represented by the value function $\mathrm{v_{\pi}(s)}$. The value function represents the value of a state $s$ under policy $\pi$ and is given by:
\begin{equation}
            \mathrm{v_{\pi}(s)}=\mathbb{E}_{\pi}\left[{\mathrm{G_{t} \mid S_{t}=s}}\right].\label{eqn:2}
\end{equation}

Here $\mathrm{G_{t}}$ is the reward sequence of the agent when in state $\mathrm{S_{t} = s}$. $\mathrm{v_{\pi}(s)}$ can also be given by the sum of all rewards $\mathrm{R_{t}}$ for each time step t where $\gamma^{k}$ represents the discounting factor:
\begin{equation}
           \mathrm{v_{\pi}(s)} =\mathbb{E}_{\pi}\left[{\mathrm{\sum_{k=0}^{\infty} \gamma^{k} R_{t+k+1} \mid S_{t}=s}}\right]. \label{eqn:3}
\end{equation}

\subsubsection{Action-value function} \label{sec: Action-value function}

We can also compute the value of performing action a, being in state s under policy $\pi$, {\it i.e.} the action-value function $\mathrm{q_{\pi}(s, a)}$. The action-valye function represents the expected return of starting from state s, performing action a and subsequently following policy $\pi$:
\begin{align}
\mathrm{q}_{\pi}(\mathrm{s}, \mathrm{a}) &=\mathbb{E}_{\pi}\left[\mathrm{G}_{\mathrm{t}} \mid \mathrm{S}_{\mathrm{t}}=\mathrm{s}, \mathrm{A}_{\mathrm{t}}=\mathrm{a}\right].\label{eqn:4}
\end{align}
We can again represent the reward sequence in terms of the sum of all rewards:
\begin{equation}
\mathrm{q}_{\pi}(\mathrm{s}, \mathrm{a})=\mathbb{E}_{\pi}\left[\sum_{\mathrm{k}=0}^{\infty} \gamma^{\mathrm{k}} \mathrm{R}_{\mathrm{t}+\mathrm{k}+1} \mid \mathrm{S}_{\mathrm{t}}=\mathrm{s}, \mathrm{A}_{\mathrm{t}}=\mathrm{a}\right]. 
\label{eqn:5}
\end{equation}

\subsection{Optimal value function and policy} \label{sec:Optimal value functions and policy}

\subsubsection{Optimal Value function} \label{sec:Optimal Value functions}

The optimal state-value function and optimal action-value function across state space $\mathcal{S}$ are then defined as $\mathrm{v_{*}(s)}$ and $\mathrm{q_{*}(s,a)}$ respectively, and are given by:
    
\begin{equation}
    \begin{array}{c}
\mathrm{v_{*}(s)}=\max _{\pi} \Big\{\mathrm{v_{\pi}(s)\Big\}}  \text { for all } \mathrm{s} \in \mathcal{S},
\end{array}\label{eqn:6}
\end{equation}

    \begin{equation}
 \begin{array}{c}
    \mathrm{q_{*}(s, a)}=\max _{\pi}  \Big\{\mathrm{q_{\pi}(s, a)\Big\}}  \text { for all } \mathrm{s} \in \mathcal{S}.
\end{array}\label{eqn:7}
\end{equation}

\subsubsection{Optimal policy} \label{sec:Optimal policy}

Using the optimal state-value function $\mathrm{v_{*}}$, actions appearing best after a one-step search will be the sub optimal ones. With optimal action-value function $\mathrm{q_{*}}$, no one-step ahead search is required from the agent. The action-value function effectively stores results of all one-step ahead searches and the agent can simply find that action maximizing action-value function $\mathrm{q_{\pi}(s, a)}$ for any given states.

\section{Methodology and Implementation} \label{sec:Methodology and Implementation}

\subsection{Environment used} \label{sec:Environment used}

The platform module, OpenGym AI \cite{opengym_ai}, containing a number of simulated environments for which RL algorithms can be compared is used in the first part of this work. We make use of the environment "Blackjack-v0", allowing one to simulate a blackjack game with the MC on-policy, MC off-policy, and the one-step q learning (hitting and standing game only). We provide the known settings of the simulated blackjack environment.

The game is played with an infinite deck. The allowable actions to the player defining the state space are to hit (hit = 1) or stand (stand = 0). After the player stops hitting or stands, the dealer reveals his face-down card and draws until his sum is greater than or equal to 17.  If the player busts, the player loses and if the dealer busts, the player wins. If neither the player nor the dealer busts, the (win, lose, draw) outcome is determined by whose sum is more favourable and closer to 21. The reward returned at each step, is 1 if the player wins, -1 if the player loses, and 0 if there is a tie. In the "Blackjack-v0" setting, we view an observation as a 3-tuple of the player's current hand total, the dealer's face up card and whether the player holds a usable ace, i.e. an ace that can be counted as an 11 without going bust. Extended environments are also implemented, allowing for more realistic assumptions to the game of blackjack. 

\subsection{Monte Carlo policy} \label{sec:Monte Carlo policy} 

We first evaluate the state-value function of a particular policy and performance of the player as a function of the number of simulations using a MC approach. Monte Carlo methods only requirement is experience, i.e. sample sequences of actions, states and rewards from an actual or simulated environment. Value estimates and policies are only changed upon the completion of an episode. To estimate the value states, we could adopt either the Monte Carlo on-policy or Monte Carlo off-policy. Any state s can be visited multiple times for the same episode. We adopt the first-visit MC, which averages returns obtained from first visits to states in each episode. to formulate the blackjack problem.

A major problem here is that many state-action pairs may never be visited and learning is thus affected as all states cannot be explored. Episodes are therefore generated with exploring starts. This means that each episode begins with a randomly chosen state and action and then follows the current policy. The agent has to therefore make sure to continue selecting states over time. We therefore adopt the MC on-policy and MC off-policy approach to learn the value states. MC on-policy attempts to evaluate or improve an existing policy while MC off-policy try to evaluate or improve one of two policies. 

\subsection{Monte Carlo improvement} \label{sec:Monte Carlo improvement}

Policy improvement is done by making the policy greedy with respect to the current value function such that: 

 \begin{equation}
    \mathrm{\pi(s)=\arg \max _{a} q(s, a)}.\label{eqn:8}
\end{equation}

\subsection{Q-Learning} \label{sec:Q-Learning}

One of the most important breakthroughs in RL was the development of an off-policy TD control algorithm known as q-Learning \cite{Sutton}. The simplest form of q-learning is a one-step q-Learning as given by: 

\begin{equation}
\begin{aligned}
\mathrm{Q_{t}}\left(\mathrm{s_{t}, a_{t}}\right) & \Leftarrow \mathrm{Q_{t}}\left(\mathrm{s_{t}, a_{t}}\right) \\
&+\alpha\left[\mathrm{R_{t}}+\gamma \max _{\mathrm{a}^{\prime}} \mathrm{Q_{t+1}}\left(\mathrm{
s_{t+1}}^{\prime}, \mathrm{a}^{\prime}\right)-\mathrm{
Q_{t+1}}\left(\mathrm{s_{t+1}}, \mathrm{a_{t}}\right)\right].
\end{aligned} \label{eqn:9}
\end{equation}

$\alpha$ is the learning rate, allowing for the determination of the update made on each time-step t and $\gamma$ is the discount rate, allowing for the determination of the value of future rewards.

The learned action-value function, Q, directly approximates $\mathrm{q_{*}}$, independent of the policy being followed which dramatically simplifies our analysis and fasten convergence. 

\subsection{Random agent} \label{sec:Random agent}
We compute the average payoff achieved by a random agent using no strategy but simply randomly hitting and standing at any state, i.e. at any round against the dealer. 

\begin{figure}[H]
	\centering
	\includegraphics[width = 80mm, scale=0.2]{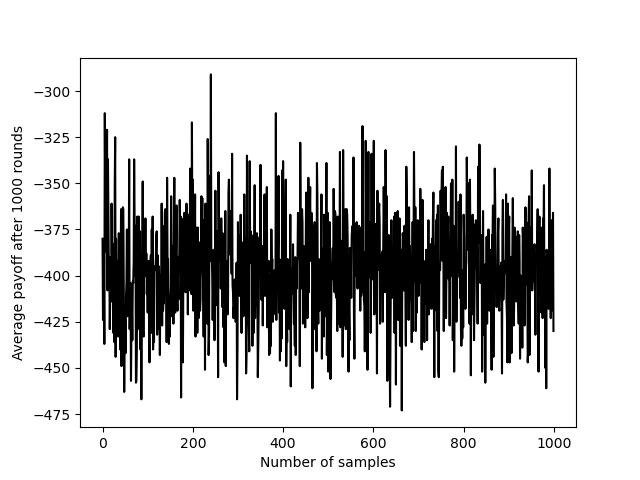}
	\caption{Here, we provide the average payoff achieved by the agent randomly hitting and standing over 1000 episodes. The figure can be recovered using python script \MYhref{https://github.com/avishburamdoyal/The-Impact-of-deck-size-Q-Learning-Blackjack/blob/main/Q-Learning/Hit\%20and\%20Stand\%20only/random_agent.py}{random\_agent.py} on GitHub resource 	\citep{github_repository}.}
	\label{fig:fig1}
\end{figure}
    
The average payoff of -396.29 achieved by the random agent is to be compared to the payoff achieved by the q-learning agent for following q-learning policy. 

\subsection{Q-Learning agent: Hit and stand only} \label{sec:Q-Learning agent: Hit and stand only}
For the base q-learning model, we implement a decaying epsilon q-learning algorithm, using the one step q-learning given by Equation \ref{eqn:9} for the blackjack environment assumed by "blackjack-v0".

To ensure an effective learning process, appropriate values for $\epsilon$. $\epsilon$ is a hyperparameter used to control the balance between exploration and exploitation in many reinforcement learning algorithms. It represents the probability of selecting a random action (exploration) versus the probability of selecting the best-known action (exploitation) at a given state. $\epsilon$ allows the agent to explore more states by forcing him to take random actions with probability $\epsilon$ \citep{Sutton}. To overcome this problem of exploration and exploitation, a decaying value of $\epsilon$ is adopted to ensure the agent minimizes exploring and exploits other states once he has learnt enough about the environment.

We initiate $\epsilon$ to be 1 and formulated it in a way that $\epsilon$ decreases to 0.9 of its initial value for the first 30\% of the training episodes, decreases to 0.2 for the next 40\% of the training rounds and reaches 0 in the next 30\% of the training rounds. The q-learning model is simulated 1000 times for 1000 rounds between the dealer and the player, using a discount factor $\gamma$ of 0.1 to keep the agent short-sighted and an arbitrary learning rate $\alpha$ of 0.05.

We perform a validation analysis of our base q-Learning model to evaluate the distribution of epsilon. 

\begin{figure} [H]
	\centering
	\includegraphics[width=0.5\textwidth]{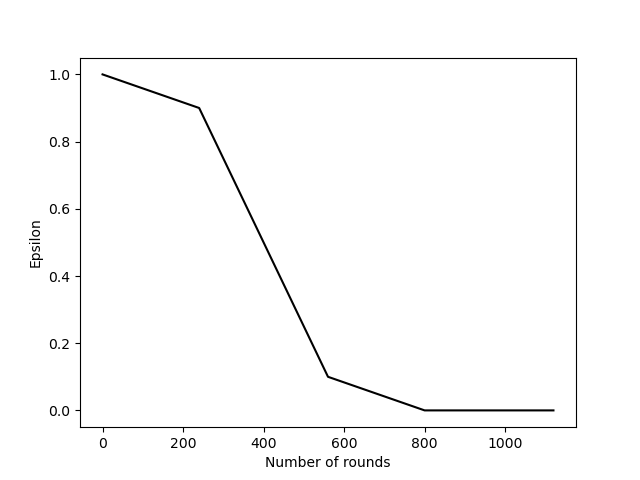}
	\caption{Here, we provide the $\epsilon$ decay plot for the $\epsilon$-greedy polcy used to generate actions of the q-learning agent. The figure can be recovered using python script \MYhref{https://github.com/avishburamdoyal/The-Impact-of-deck-size-Q-Learning-Blackjack/blob/main/Q-Learning/Hit\%20and\%20Stand\%20only/validation_analysis.py}{validation\_analysis.py} on GitHub resource \citep{github_repository}.} 
	\label{fig:fig2}
\end{figure}
    
Figure \ref{fig:fig2} demonstrates the rate of decay of epsilon as the number of rounds is increased. Once $\epsilon$ reaches a tolerance value, which happens around a total of 800 rounds, the agent has explored enough and the value of $\epsilon$ converges to 0.


\begin{figure} [H]
 	\centering
    	\includegraphics[width=0.5\textwidth]{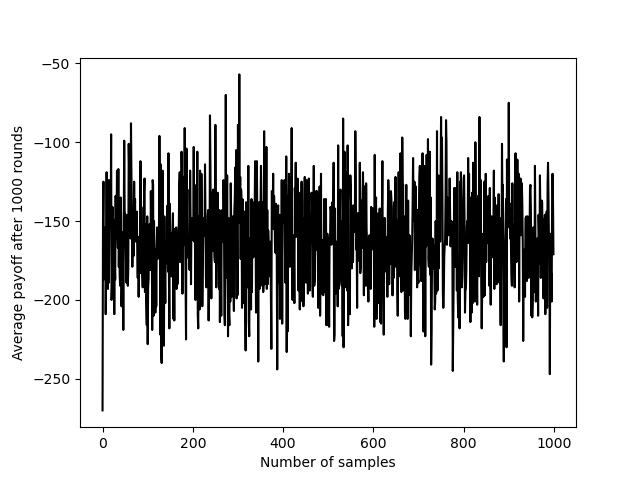} 
    	\caption{Here, we provide the average payoff achieved by the q-learning agent over 1000 episodes. The figure can be recovered using python script \MYhref{https://github.com/avishburamdoyal/The-Impact-of-deck-size-Q-Learning-Blackjack/blob/main/Q-Learning/Hit\%20and\%20Stand\%20only/q_agent.py}{q\_agent.py} on GitHub resource \citep{github_repository}.}  
	\label{fig:fig3}
\end{figure}

The trained agent, using the chosen hyperparameters achieves an average payoff of -127.677, representing an increase in average payoff of 62.74\% to the random agent. This is indicative, of a better and more rewarding policy learnt by the agent following a decaying epsilon q-learning policy.

\begin{figure} [H]
	\centering
	\subfloat[No usable ace]{\label{subfig:1stsiman}
	\includegraphics[width = 80mm, scale=0.8]{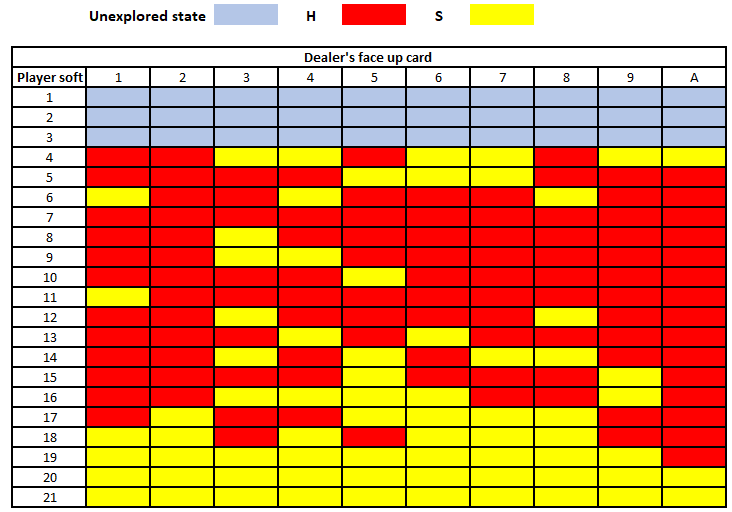}}\\
	\subfloat[Usable ace]{\label{subfig:2ndsiman}
	\includegraphics[width = 80mm, scale=0.8]{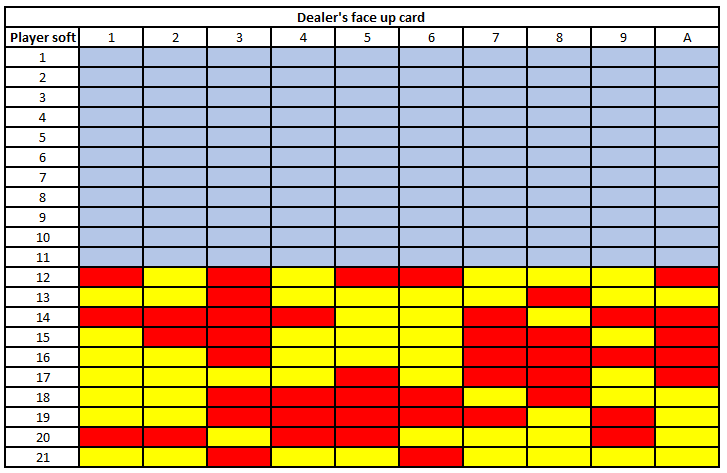}}
	\caption{The strategy learnt by the q-learning agent can be recovered using python script \MYhref{https://github.com/avishburamdoyal/The-Impact-of-deck-size-Q-Learning-Blackjack/blob/main/Q-Learning/Hit\%20and\%20Stand\%20only/strategy_learnt.py}{strategy\_learnt.py} on GitHub resource \citep{github_repository}.}
	\label{fig:fig4}
\end{figure}

It is also particularly interesting to analyze the strategy chart learnt by the agent. Figure \ref{fig:fig4} is a tabular representation of the strategy learned by the agent. The left part shows the strategy for which the player has no usable ace and the right half indicates the strategy for which the player has a usable ace. There are 3 cases where the q-learning agent either hits (H), stands (S) or doesn't reach a state.

In the second part of of Figure \ref{fig:fig4}, 110 states are not explored, given the player has a usable ace, meaning the player's hand total will always be more than 11.

\subsection{Q-Learning policy (all actions implemented correctly)} \label{sec:Q-Learning policy (all actions implemented correctly)}

Two further extended q-Learning models are implemented making the game settings of blackjack more realistic. We assess the impact that a variation in deck size may have on the learning ability of the agent. In addition, we analyze the strategy learnt by the q-learning agent.

\subsubsection{Impact of deck-size Q-Learning Blackjack} \label{sec:Impact of deck-size Q-Learning Blackjack}

We analyze whether a variation in deck size influences the learning rate of the q-learning agent for a game between one player and the dealer. We assume a standard blackjack game allowing for all actions: hitting, standing, doubling down, splitting and surrendering. The metric used to analyze the learning rate of the agent is winning odds \% against the dealer for using the q-learning strategy. The one-step q-learning as given by Equation \ref{eqn:9} in Section \ref{sec:Q-Learning} is adapted. 

We compare the performance of the trained agent over 4 to 8 decks played across three different counting systems. We choose an arbitrary learning rate $\alpha$ of 0.05 and a discount factor $\gamma$ of 0.1. We train the agent for 500000 episodes and backtest the game 50000 times based on training results.

\begin{table}[H]
\centering
\begin{tabular}{|c|c|c|c|}
\hline \textbf{Deck size} & \textbf{Hi-Lo} & \textbf{Zen count} & \textbf{Uston APC} \\
\hline 4 & 42.24\% & 41.83\% & 41.45\% \\
\hline 5 & 42.59\% & 41.88\% & 41.43\% \\
\hline 6 & 42.01\% & 41.96\% & 42.21\% \\
\hline 7 & 42.14\% & 42.08\% & 41.90\% \\
\hline 8 & 41.13\% & 41.91\% & 42.11\% \\
\hline
\end{tabular}
\caption{The winning odds \% can be recovered using python script \MYhref{https://github.com/avishburamdoyal/The-Impact-of-deck-size-Q-Learning-Blackjack/blob/main/Q-Learning/Impact\%20of\%20deck\%20size/main.py}{main.py} for which the variation in counting system is implemented from python script \MYhref{https://github.com/avishburamdoyal/The-Impact-of-deck-size-Q-Learning-Blackjack/blob/main/Q-Learning/Impact\%20of\%20deck\%20size/env_state.py} {env\_state.py} and variation in deck size from Python script \MYhref{https://github.com/avishburamdoyal/The-Impact-of-deck-size-Q-Learning-Blackjack/blob/main/Q-Learning/Impact\%20of\%20deck\%20size/deck_state.py} {deck\_state.py} on GitHub resource \citep{github_repository}.}
\label{tab:count5}
\end{table}

We observe that for a blackjack game between a single player and a dealer, played with 4 to 8 decks, the discrepancy in winning odds \% across varying deck size is not significant across the 3 counting systems: Hi-Lo, Zen count and Uston APC. The Hi-Lo system might seem to be the most rewarding system for this setting. 

We evaluate the learning performance of the trained agent for the 3 counting systems across varied deck size, beyond the standard 4-8 decks used in casinos.  

 \begin{figure} [H]
\begin{center}
    \includegraphics[width=0.5\textwidth]{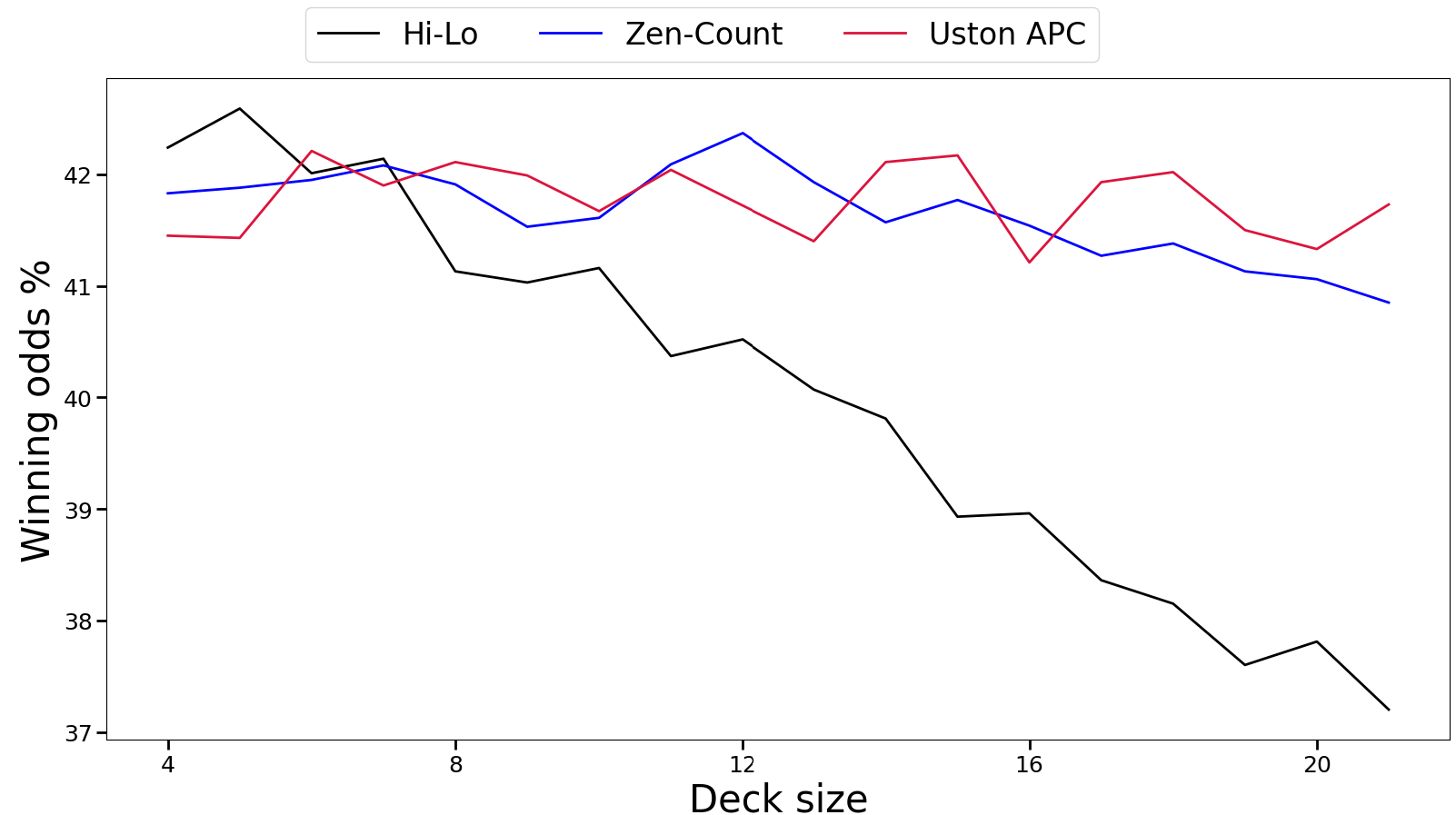}
    \caption{Figure \ref{fig:fig5} can be recovered using python script \MYhref{https://github.com/avishburamdoyal/The-Impact-of-deck-size-Q-Learning-Blackjack/blob/main/Q-Learning/Impact\%20of\%20deck\%20size/Impact\%20of\%20deck\%20size\%20Q-Learning.py}{Impact of deck size Q-Learning.py}, to which all average payoffs were calculated using, \MYhref{https://github.com/avishburamdoyal/The-Impact-of-deck-size-Q-Learning-Blackjack/blob/main/Q-Learning/Impact\%20of\%20deck\%20size/main.py}{main.py} for which the deck size and counting rule variation are implemented using python scripts \MYhref{https://github.com/avishburamdoyal/The-Impact-of-deck-size-Q-Learning-Blackjack/blob/main/Q-Learning/Impact\%20of\%20deck\%20size/deck_state.py}{deck\_state.py} and \MYhref{https://github.com/avishburamdoyal/The-Impact-of-deck-size-Q-Learning-Blackjack/blob/main/Q-Learning/Impact\%20of\%20deck\%20size/env_state.py}{env\_state.py} respectively, available on GitHub resource \citep{github_repository}.}
    \label{fig:fig5}
    \end{center}
\end{figure}

From Figure \ref{fig:fig5} we observe little discrepancy in terms of winning odds \% across deck size (4-21) under the Zen Count and Uston APC system and a steeper downward trend of the winning odds \% of the trained agent for using the Hi-Lo method.

We conduct a comparative analysis of the winning odds \% between the Zen-count and Uston APC methods, with a particular emphasis on a game with more than 21 decks. Empirical data reveals that both Zen-count and Uston APC consistently demonstrate superior performance compared to the widely-used Hi-Lo method. Additionally, we observe a significant downward trend in the Hi-Lo method’s winning odds\% as the number of decks increases, thus highlighting the need to explore more effective counting strategies in games with a larger deck count. Using multiple decks also makes it more challenging to predict the probability of certain cards being dealt, thus increasing the house edge.

\begin{figure} [H]
\begin{center}
    \includegraphics[width=0.5\textwidth]{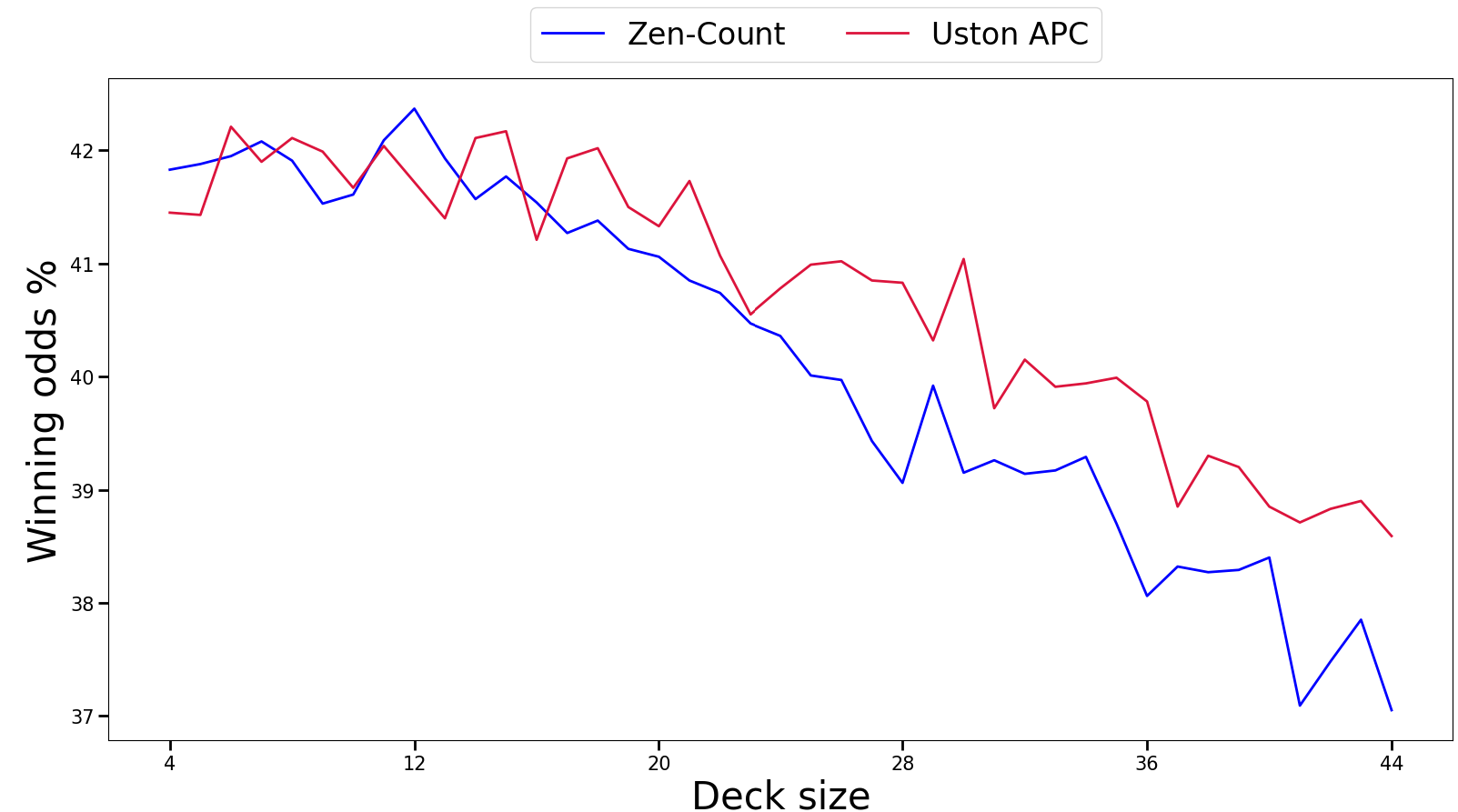}  
    \caption{Figure \ref{fig:fig6} can be recovered using python script \MYhref{https://github.com/avishburamdoyal/The-Impact-of-deck-size-Q-Learning-Blackjack/blob/main/Q-Learning/Impact\%20of\%20deck\%20size/Impact\%20of\%20deck\%20size\%20Q-Learning.py}{Impact of deck size Q-Learning.py}, to which all average payoffs are calculated using, \MYhref{https://github.com/avishburamdoyal/The-Impact-of-deck-size-Q-Learning-Blackjack/blob/main/Q-Learning/Impact\%20of\%20deck\%20size/main.py}{main.py} for which the deck size and counting rule variation are implemented using python scripts \MYhref{https://github.com/avishburamdoyal/The-Impact-of-deck-size-Q-Learning-Blackjack/blob/main/Q-Learning/Impact\%20of\%20deck\%20size/deck_state.py}{deck\_state.py} and \MYhref{https://github.com/avishburamdoyal/The-Impact-of-deck-size-Q-Learning-Blackjack/blob/main/Q-Learning/Impact\%20of\%20deck\%20size/env_state.py}{env\_state.py} respectively, available on GitHub resource \citep{github_repository}.} 
    \label{fig:fig6}
    \end{center}
\end{figure}

\subsubsection{Policy learnt} \label{sec:Policy learnt}
\* 

The strategy table for the extended q-learning model\footnote{We use an epsilon greedy q-learning algorithm with a learning rate $\alpha$ of 0.01, a discount factor $\gamma$ of 0.1 and allow the dealer to hit on soft 17 across 10000000 blackjack games simulated.} is also analyzed. 

\begin{figure} [H]
	\centering
	\subfloat[Player hard total]{\label{subfig:correct}
	\includegraphics[width = 80mm, scale=0.8]{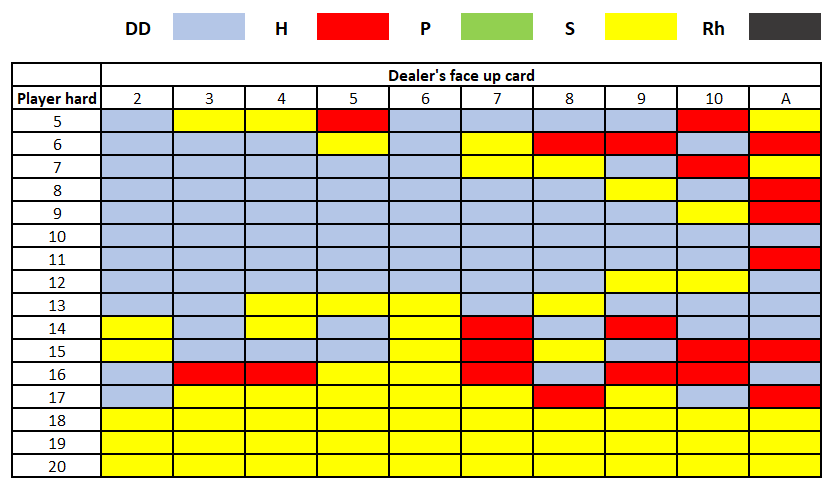}} \\
	\subfloat[Player soft total]{\label{subfig:notwhitelight}
	\includegraphics[width = 80mm, scale=0.8]{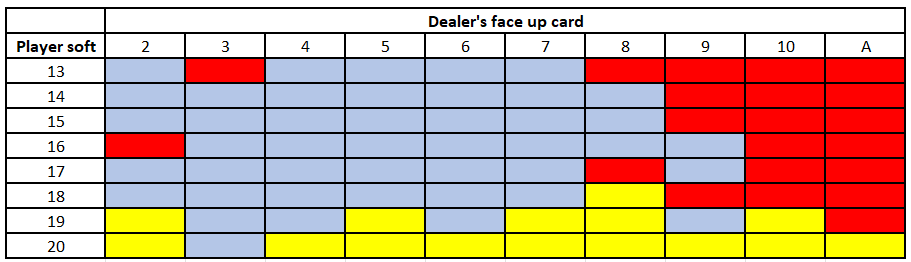}}\\ 
	\subfloat[Player splits]{\label{subfig:nonkohler}
	\includegraphics[width = 80mm, scale=0.8]{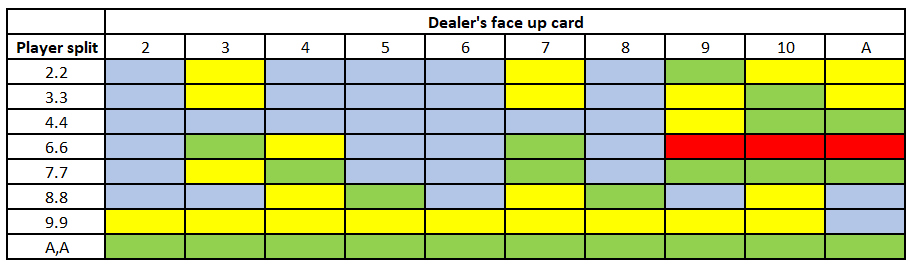}} 
	\caption{The policy learnt is generated in a csv format  using python script \MYhref{https://github.com/avishburamdoyal/The-Impact-of-deck-size-Q-Learning-Blackjack/blob/main/Q-Learning/Analyzing\%20Strategy\%20Table\%20(All\%20actions\%20implemented)/Q\%20Learning\%20-\%20states\%20mapping.py}{Q Learning - states mapping.py} after the Q-scores are calculated using python script \MYhref{https://github.com/avishburamdoyal/The-Impact-of-deck-size-Q-Learning-Blackjack/blob/main/Q-Learning/Analyzing\%20Strategy\%20Table\%20(All\%20actions\%20implemented)/Q\%20Learning\%20-\%20qscores.py}{Q Learning - qscores.py} on GitHub resource \citep{github_repository}.}
	\label{fig:fig7}
\end{figure}

Comparing the strategy learnt for which the player may have hard totals, soft totals and pairs, we observe similarities to the strategy table adapted from \citep{mich} in Section \ref{sec:Strategy Tables} of the Appendix in 64.78\% of the cases. This is a satisfactory result but we note that the agent has not learnt when to surrender even the action is allowable.

\subsection{Bringing down the house} \label{sec:Bringing down the house} 

ect thWe implement a blackjack simulator allowing a player to perfectly use the basic strategy table, outlined in Section \ref{sec:Strategy Tables} of the Appendix and the Hi-Lo system in Section \ref{sec:Ten count and Hi-lo system}. This player is the card counter. We compare the performance rates of the card counter to that of other players (called random agents) taking randomly intuitive actions. 

The game starts off with a fixed number of decks. After each round, each player places an initial bet. The dealer then deals two cards to each player and himself. Each player chooses their actions followed by the dealer choosing his. Based on favourable scores, money between the dealer and each player is exchanged. This is when one round ends. Once the number of decks used is nearly depleted, there are not enough cards to be dealt to allow play to happen. This is when one simulation ends. A second simulation begins when a fresh number of decks is again placed on the table after which another game with multiple rounds resume. A simulation comes to a halt when we have less than or equal to 25 cards. 

We allow the use multiple decks and simulate the game for 4-8 decks. We allow for a multiplayer game and simulate the game for 1-7 players. Actions are based off the strategy table, indicated in Section \ref{sec:Available actions} with the exception of insurance. We impose an intuitive condition where the random agent hits 3 times for a hand total less than 6 and hit 2 times otherwise. The choice of hitting 2 times is chosen minimizing the risk of player busting. A condition where the player splits for a hand total of greater or equal to 4 and less than 18 is also imposed. The card counter uses the Hi-Low method and the other players randomly decide on the \% of their bankroll to bet for each round. We allow the dealer the choice to either stand or hit on having a soft 17.  We set a fixed bankroll for all players and a fixed one (proportionately higher) for the dealer. It is assumed that, when the dealer goes bankrupt, the house still pays the winning agent and goes negative. The dealer and players all stays in the game and their bankroll stay updated with respect to positive and negative rewards and payments. 

\textbf{Case 1}: 4 Players, 6 Decks, Dealer stands on soft 17 and 10000 simulations

\begin{table}[H]
\centering
\begin{tabular}{|c|c|c|c|c|}
\hline
\textbf{Players} & \textbf{Win \%}  &  \textbf{Draw \%} & \textbf{Loss \%} & \textbf{Bankroll}\\
\hline
Card counter & 14.55 \% & 1.35 \% & 3.21 \% & R14, 647, 394 \\
\hline
Random agent 1 & 11.03 \% & 1.43 \% & 7.52 \% & -R6, 835, 650 \\
\hline
Random agent 2 & 11.01 \% & 1.44 \% & 7.52 \% & -R6, 989, 950\\
\hline
Random agent 3 & 11.02 \%  & 1.47 \% & 7.42 \% & -R6, 949, 200\\
\hline
Dealer & 3.21 \% & 1.35 \% & 14.55 \% & -R14, 622, 385 \\
\hline
\end{tabular}
\caption{The win, draw, loss \% and bankroll balance can be recovered using Python script \MYhref{https://github.com/avishburamdoyal/The-Impact-of-deck-size-Q-Learning-Blackjack/blob/main/Basic\%20Strategy\%20\%2B\%20Hi-Lo/Python/Strategy\%20Table\%20\%2B\%20Hi-low.py}{Strategy Table + Hi-low.py} on GitHub resource \citep{github_repository}.} 
\label{tab:count6}
\end{table}

\textbf{Case 2}: 6 Players, 8 Decks, Dealer hits on soft 17 and 10000 simulations

\begin{table}[H]
\centering
\begin{tabular}{|c|c|c|c|c|}
\hline
\textbf{Players} & \textbf{Win \%}  &  \textbf{Draw \%} & \textbf{Loss \%} & \textbf{Bankroll}\\
\hline
Card counter & 8.58 \% & 0.81 \% & 2.06 \% & R14, 061, 791 \\
\hline
Random agent 1 & 6.58 \% & 0.84 \% & 4.53 \% & -R6, 975, 850 \\
\hline
Random agent 2 & 6.56 \% & 0.87 \% & 4.53 \% & -R7, 012, 000\\
\hline
Random agent 3 & 6.56 \%  & 0.85 \% & 4.54 \% & -R6, 980, 700\\
\hline
Random agent 4 & 6.56 \% & 0.85 \% & 4.55 \% & -R7, 177, 450\\
\hline
Random agent 5 & 6.61 \% & 0.88 \% & 4.47 \% & -R6, 961, 650\\
\hline
Dealer & 2.06 \% & 0.81 \% & 8.58 \% & -R18, 691, 233 \\
\hline
\end{tabular}
\caption{The win, draw, loss \% and bankroll balance can be recovered using Python script \MYhref{https://github.com/avishburamdoyal/The-Impact-of-deck-size-Q-Learning-Blackjack/blob/main/Basic\%20Strategy\%20\%2B\%20Hi-Lo/Python/Strategy\%20Table\%20\%2B\%20Hi-low.py}{Strategy Table + Hi-low.py}on GitHub resource \citep{github_repository}.} \label{tab:count7}
\end{table}

Tables \ref{tab:count6} and \ref{tab:count7} indicate the outperformance of the card counter against the dealer in terms of win \% relative to the random agents whether the dealer hits or stands on soft a soft 17 hand. We also observe a large discrepancy between the bankroll of the card counter compared to the other players and dealer. Even for an initial bankroll ratio of 1:200 in favour of the house, the dealer goes bankrupt for having a card counter perfectly using the basic strategy table and the Hi-Lo system.

Next, we investigate the performance of the card counter relative to another random agent against the dealer as a function of simulation size, number of players and deck size.

\subsubsection{Performance across simulation size} \label{sec:Performance across simulation size}
\*

\begin{figure} [H]
	\centering
	\subfloat[Simulation: 500, Players: 4, Decks: 6, Dealer: Stands on soft 17]{\label{subfig:1stsim}
	\includegraphics[width=0.5\textwidth]{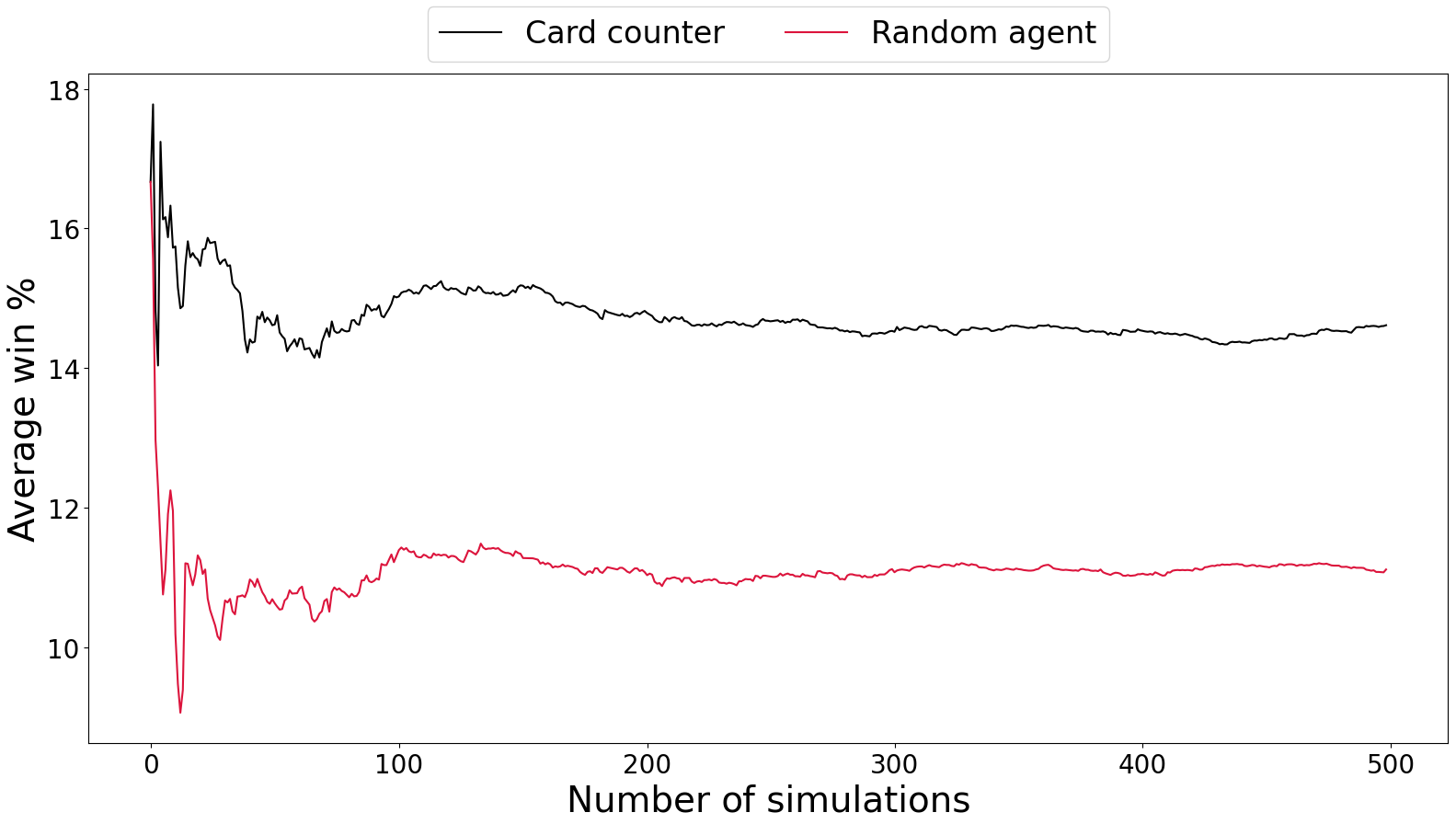}}\\
	\subfloat[Simulation: 10000, Players: 6, Decks: 8, Dealer: Hits on soft 17]{\label{subfig:2ndsim}
	\includegraphics[width=0.5\textwidth]{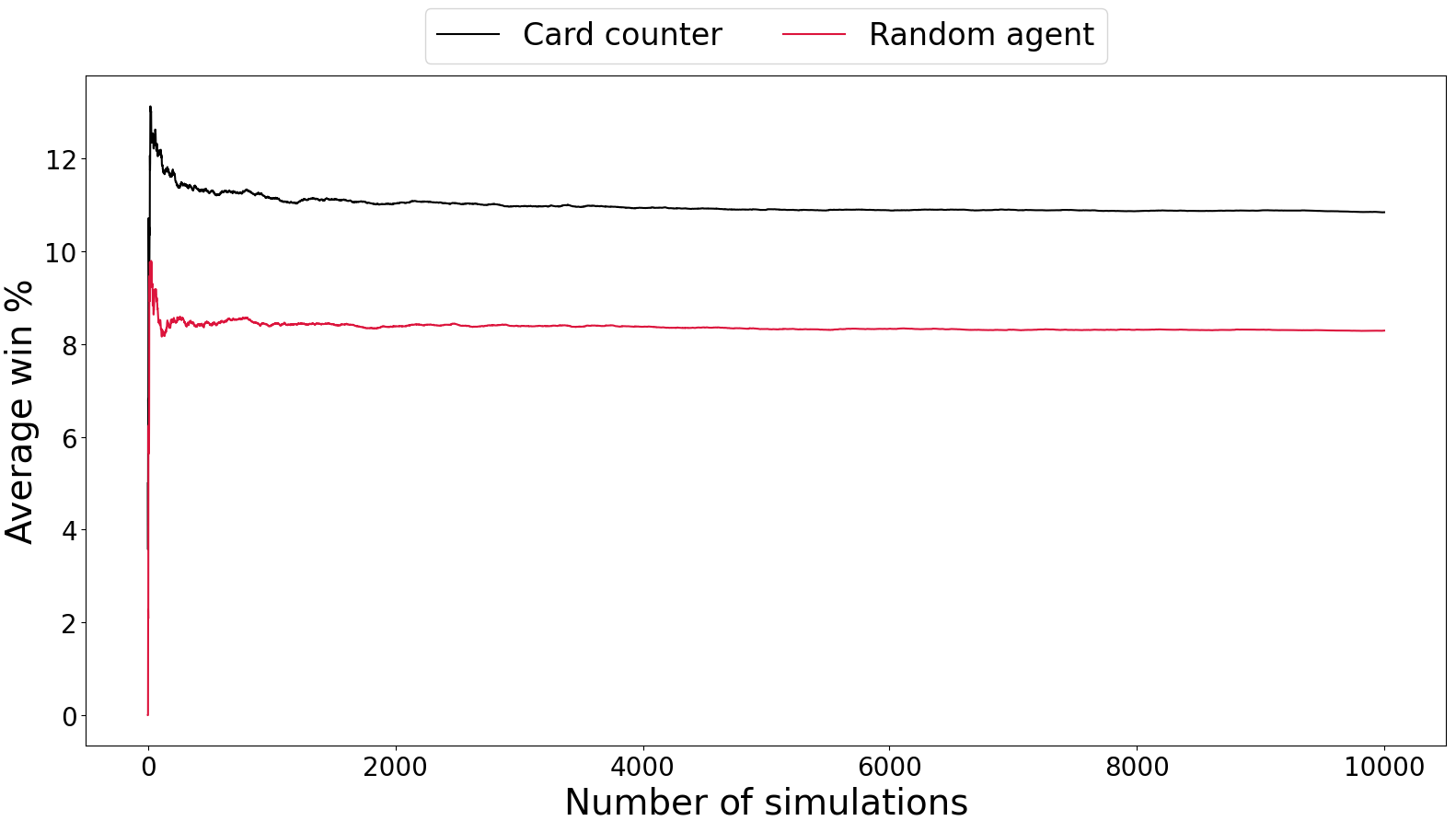}}
	\caption{Figure \ref{fig:fig8} can be recovered using 	Python script  \MYhref{https://github.com/avishburamdoyal/The-Impact-of-deck-size-Q-Learning-Blackjack/blob/main/Basic\%20Strategy\%20\%2B\%20Hi-Lo/Python/Performance\%20vs\%20simulations.py}{Performance vs simulations.py} on GitHub resource \citep{github_repository}. }
	\label{fig:fig8}
\end{figure}

Figure \ref{fig:fig8} shows a clear outperformance of the card counter to the random agent against the dealer across both variations in the number of players, deck size and whether the dealer hits or stands on a soft 17 hand.

\subsubsection{Performance across number of players} \label{sec:Performance across number of players}
\*

\begin{figure} [H]
	\centering 
	\includegraphics[width=0.5\textwidth]{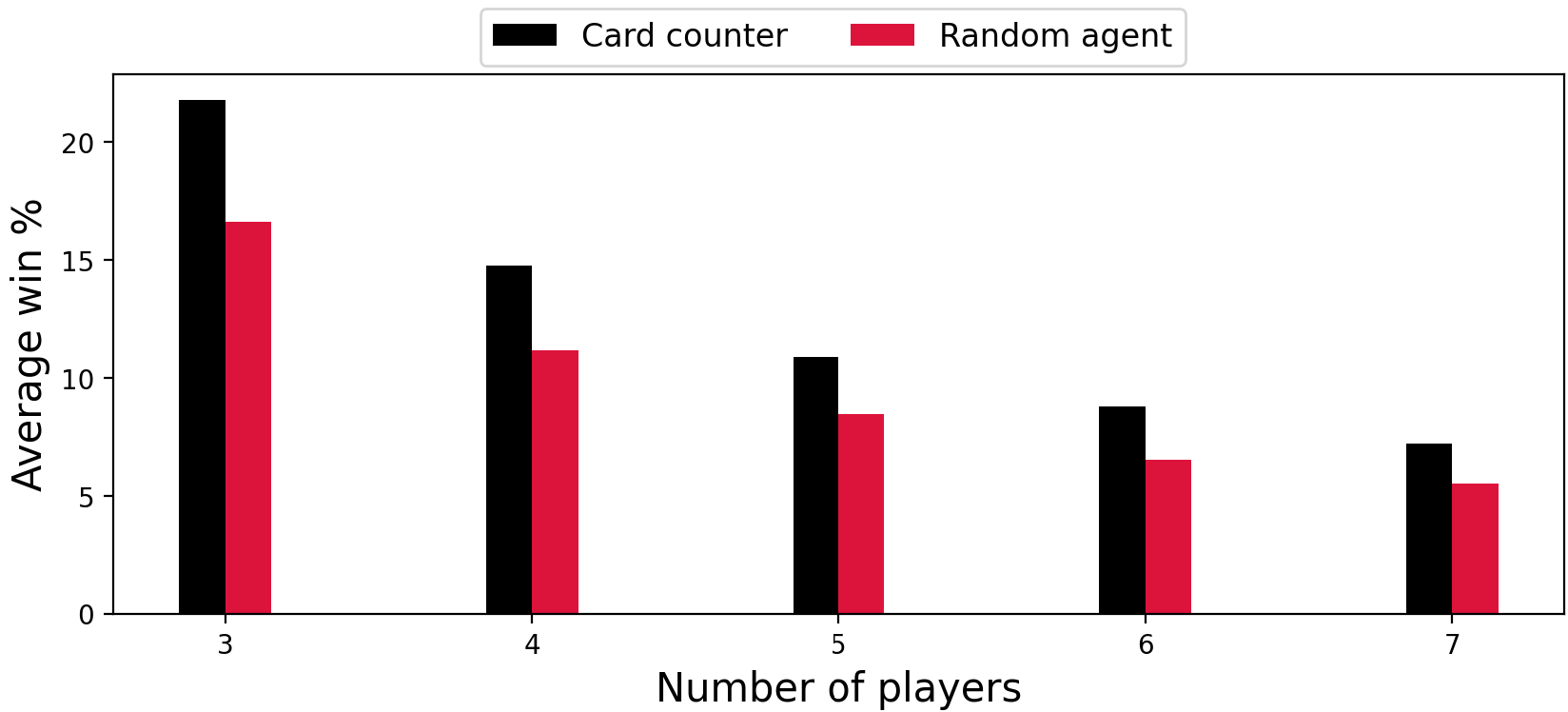}
    	\caption{Simulation: 2000, Players: 3-7, Decks: 8, Dealer: Stands on soft 17. Figure \ref{fig:fig9} can be recovered using 	python script \MYhref{https://github.com/avishburamdoyal/The-Impact-of-deck-size-Q-Learning-Blackjack/blob/main/Basic\%20Strategy\%20\%2B\%20Hi-Lo/Python/Performance\%20vs\%20number\%20of\%20players.py} {Performance vs number of players.py} on GitHub resource \citep{github_repository}. } 
	\label{fig:fig9}
\end{figure}

Figure \ref{fig:fig9}  also indicates the outperformance of the card counter to the random agent across increased number of players and varied deck size.

\subsubsection{Performance across deck size} \label{sec:Performance across deck size}
\* 

\begin{figure} [H]
    \includegraphics[width=0.5\textwidth]{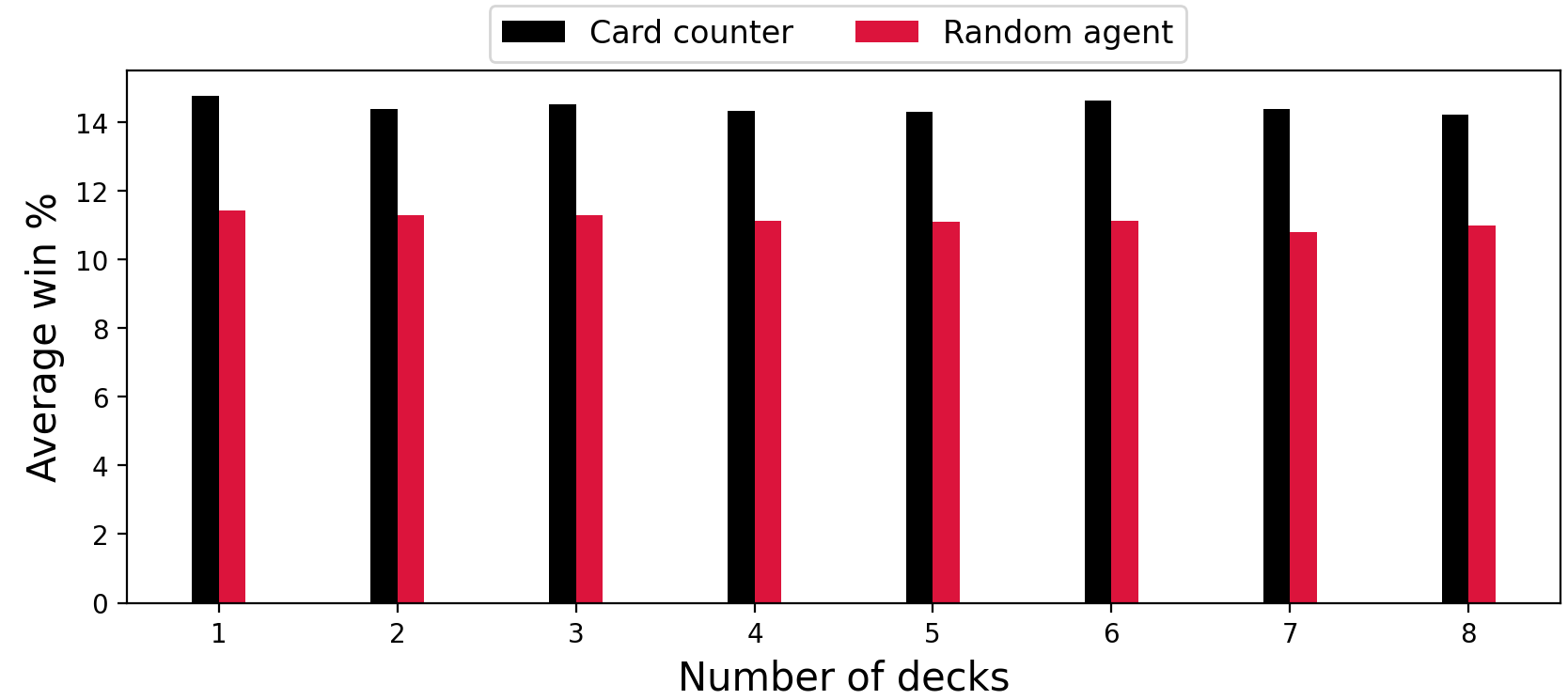}
    \caption{Simulation: 1000, Players: 6, Decks: 1-8, Dealer: Hits on soft 17. Figure \ref{fig:fig10} can be recovered using 	python script \MYhref{https://github.com/avishburamdoyal/The-Impact-of-deck-size-Q-Learning-Blackjack/blob/main/Basic\%20Strategy\%20\%2B\%20Hi-Lo/Python/Performance\%20vs\%20deck\%20size.py} {Performance vs deck size.py} on GitHub resource \citep{github_repository}. } 
	\label{fig:fig10}
\end{figure}

Figure \ref{fig:fig10} indicates that across all variations in the number of players, simulation size and whether the dealer stands or hits on soft 17, the card counter always outperforms the random agent. The number of decks in play do not seem to influence the discrepancy in win\% against the dealer between the card counter and the random agent.

\section{Conclusion} \label{sec:Conclusion}

This paper set out to better  understand the performance of the decaying epsilon greedy algorithm to blackjack and whether variations in blackjack settings may affect performance of the q-learning agent, particularly across a variation in deck size.. In particular when the learning agent is faced with combinations of increasing deck size, that represents the available states faced by the learning algorithm, and the number of players. The change in number of players moves the game into a simple multi-agent game. Here it is still found that a traditional q-learning approach is preferred relative to the other reinforcement learning algorithms that were implemented. The reasons seems to be that this still allows for realistically learning to happen through the use of the decaying epsilon greedy parameter while requiring fewer simulations for reasonable convergence to the right policy with improved payoffs.

We demonstrate that for deck sizes 4 to 8, learning blackjack using a decaying epsilon greedy q-learning approach, and adopting one of: Hi-Lo system, Zen count or Uston APC to count cards, does not significantly affect learning performance of the agent. As the deck size is extended to a maximum of 21 decks, we observe that across all deck sizes, the agent's average winning odds \% remain roughly the same and no apparent trend is noticed for the Zen count and Uston APC. Under the Hi-Lo method however, extending the deck size to a maximum of 21 clearly indicates a steeper downward trend in winning odds \% of the learning agent. We would therefore can recommend the use of either the Zen count or Uston APC -- two more seemingly rewarding counting systems.

It has also been shown that the decaying epsilon greedy q-learning provides satisfactory results in learning the optimal policy {\it i.e.} the optimal strategy table, but note that the agent never learns to surrender.

We also investigate the performance of a player perfectly sizing his bet using the Hi-Lo system, and perfectly requesting actions following the basic strategy table as outlined in Section \ref{sec:Appendix} of the Appendix. It has been found that the card counter always outperforms the house irrespective of the deck size, number of players or simulations. It is, however, observed that across an increased number of players, the average winnings \% of both the card counter and other players decrease exponentially. This unsurprisingly suggests that the algorithm should play on a table with fewer players. Increasing the deck size for a non q-learning process should not significantly affect the winning \% of the card counter when using the Hi-lo system. In future work we could move away from the 7 player limit to try find out when learning is saturated as one moves towards an infinite player and decks size problem. 

\section*{Acknowledgment} \label{sec:Acknowledgment}
We would like to thank Matt Dicks for his comments and the Department of Statistical Sciences and in particular Francesca Little and Etienne Piennar.
\onecolumn

    \section{Appendix} \label{sec:Appendix}
    
    \subsection{Strategy Table} \label{sec:Strategy Tables}
    
    \begin{figure}[H]
    	\centering
        	\includegraphics[width = 140mm, scale=1.0]{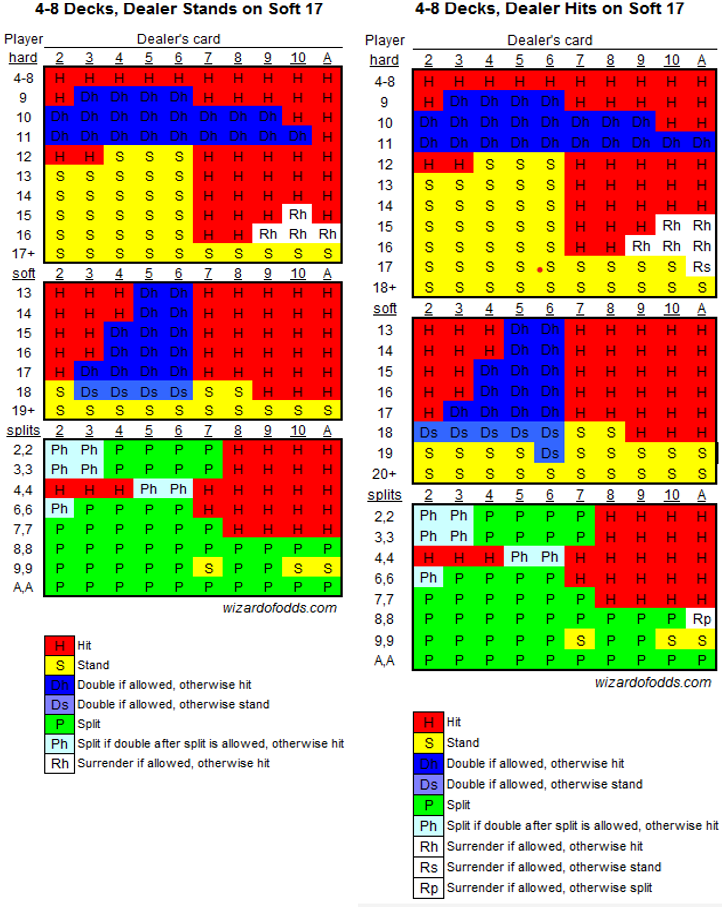}
                \captionsetup{width=\textwidth}
        	\caption{The above two tables are the strategy tables adapted from \cite{mich}. To use the basic strategy, a player looks  up  his  hand  along  the  left  vertical  edge  and  the 	dealer’s up card along the top.  In both cases an A stands for ace.  From top to bottom are the hard totals, soft totals, and splittable hands. There are two charts depending on whether the dealer hits or stands on soft 17.} 
    	\label{fig:fig11}
    \end{figure}

\twocolumn

\subsection{Algorithm} \label{sec: Algorithm}

The q-Learning algorithm procedural form for a decaying greedy epsilon policy case is shown below:

\begin{tcolorbox}[
    standard jigsaw,
    title=Q-Learning off-policy TD control algorithm,
    opacityback=0, 
]
\textbf{Initialize $\mathrm{Q(s, a)}, \forall \mathrm{s} \in \mathcal{S}, \mathrm{a} \in \mathcal{A}(\mathrm{s}),$ arbitrarily, and $\mathrm{Q}($ terminal-state}, $\cdot)=0$
\vspace{0.5cm}

\textbf{Repeat (for each episode)}:\\
\phantom{x}\hspace{2ex}Initialize $\mathrm{S}$\\ 
\phantom{x}\hspace{2ex}Repeat (for each step of episode):\\
\phantom{x}\hspace{4ex}Choose A from S using policy derived from Q\\ 
\phantom{x}\hspace{4ex}Take action A, observe R, $\mathrm{S}^{\prime}$\\
\phantom{x}\hspace{4ex}
$\begin{array}{l}
\text{Q}\left(\text{S}_{\text{t}}, \text{A}_{\text{t}}\right) \leftarrow \text{Q}\left(\text{S}_{\text{t}}, \text{A}_{\text{t}}\right)+ \\
\alpha\left[\text{R}_{{\text{t}}+1}+\gamma \max _{\text{a}} \text{Q}\left(\text{S}_{{\text{t}}+1}, \text{a}\right)-\text{Q}\left(\text{S}_{\text{t}}, \text{A}_{\text{t}}\right)\right]
\end{array}$
\phantom{x}\hspace{2ex}until S is terminal
\end{tcolorbox}
\textbf{Algorithm}: \begin{footnotesize} Here we provide the procedural form of the one-step q-learning, adapted from \citep{Sutton}. This algorithm is fundamental to the project for purpose of analyzing the strategy learnt, average payoff achieved by a trained agent and the impact that a variation in deck size might have on learning performance. The Python implementation of the one-step q-learning can be found in  \MYhref{https://github.com/avb1597/The-Impact-of-deck-size-Q-Learning-Blackjack/tree/main/Q-Learning/Hit\%20and\%20Stand\%20only}{Qbase} on Github resource \citep{github_repository}.  
\end{footnotesize}

\subsection{Modules for extended Q-Learning model} \label{sec:Modules for extended Q-Learning model}

\textbf{Module set 1: Impact of deck size learning}:

The Python implementation in assessing the impact of deck size is done using 5 modules: \texttt{deck\_state}, \texttt{env\_state}, \texttt{training}, \texttt{testing} and \texttt{main}.

\begin{enumerate} 
    \item \texttt{deck\_state}: A class \texttt{Deck} for which functions allowing for: a variation in the number of decks, shuffling of pack, drawing of cards and returning the number of remaining cards in deck, to be used to size bets, are defined. 
    \item \texttt{env\_state}: A class \texttt{handState}. is defined. We make the distinction between the dealer's and the player's hand. Distinction about "special type hands": hard hands, soft hands and pair hands are also considered. The state of the environment is definted as a tuple of the player's hand total and the dealer's face up card.  A function \texttt{update\_deck\_state} is created to update the deck state based off card counting and a function \texttt{update\_card\_state} which updates the card states (total) upon actions being requested.
    \item \texttt{training}: A class \texttt{Train\_agent} is defined, allowing an agent to be trained for particular settings. The allowable actions includes: hitting, standing, doubling down and splitting. A function \texttt{row\_index} is created which returns the row index of the Q-matrix for a given card state. Two additional functions, \texttt{stand\_reward} and \texttt{doubledown\_reward} are also created to return immediate rewards for a player either standing (no reward) and for a player doubling down (getting double his initial bet for placing a bet of equal size in between game). The rewards are modelled as $0, 1$ and 2 for the player losing, drawing and winning respectively against the dealer for standing in a round. Also, rewards are also modelled as $-1, 1$ and 3 for the player losing, drawing and winning against the dealer for doubling down in a round.
    \item \texttt{blackjack\_game}: The \texttt{training} module is created which takes as argument the \texttt{card\_state}, i.e. the player's total and dealer's face up card, allowing for the blackjack play and returns profit made by the learning agent. The agent is then trained for a specified number of times. We note that a new card set is presented to the table when the remaining cards in the deck is less than 30. 
    \item \texttt{testing}: A further module adapted from \citep{impact_of_decksize}. Under \texttt{testing}, we have a class \texttt{Backtest\_rl\_agent} taking into account the game settings adapted for the training phase. An additional function, \texttt{backtest\_model} is then created which returns profit made by the learning agent by backtesting the game based off the training results.
 \end{enumerate}
\vspace{1.72cm}

\textbf{Module set 2: Strategy chart}:

In this case, we work with fewer modules. More specifically, we have 3 modules: \textbf{main}, \textbf{Q Learning - qscores} and \textbf{Q Learning - states mapping}. 

\begin{enumerate}
    \item \textbf{main}: Represents a blackjack simulator allowing for play to happen between a dealer and one player for which an infinite deck/deck with replacement is assumed on each round. 
    \item \textbf{Q Learning}: Allows the agent/player to q-learn the blackjack game, as explained in Section \ref{sec:Q-Learning} and returns a csv file: \textbf{Qsa\_values} of the q-values to each states.
    \item \textbf{Qsa\_values}: Maps states of the strategy table to the respective actions based on the q-values returned by module \textbf{Q Learning - qscores}.
\end{enumerate}
\newpage 

\bibliographystyle{IEEEtranN}
\bibliography{conference_041818}

\end{document}